# MTCNET: MULTI-TASK LEARNING PARADIGM FOR CROWD COUNT ESTIMATION

*Abhay Kumar*[*]  *Nishant Jain*[*]  *Suraj Tripathi*[♦]  *Chirag Singh*[♦]  *Kamal Krishna*

Samsung R&D Institute- India, Bangalore

## ABSTRACT

We propose a Multi-Task Learning (MTL) paradigm based deep neural network architecture, called MTCNet (Multi-Task Crowd Network) for crowd density and count estimation. Crowd count estimation is challenging due to the non-uniform scale variations and the arbitrary perspective of an individual image. The proposed model has two related tasks, with Crowd Density Estimation as the main task and Crowd-Count Group Classification as the auxiliary task. The auxiliary task helps in capturing the relevant scale-related information to improve the performance of the main task. The main task model comprises two blocks: VGG-16 front-end for feature extraction and a dilated Convolutional Neural Network for density map generation. The auxiliary task model shares the same front-end as the main task, followed by a CNN classifier. Our proposed network achieves 5.8% and 14.9% lower Mean Absolute Error (MAE) than the state-of-the-art methods on ShanghaiTech dataset without using any data augmentation. Our model also outperforms with 10.5% lower MAE on UCF_CC_50 dataset.

## 1. INTRODUCTION

Recently, there has been an accelerated growth in research on crowd understanding due to its varied applications including crowd management, public safety, video surveillance, crowd flow analysis etc. It could help to prevent traffic congestion and stampedes at crowded events. In the last few years, researchers have worked on various crowded scene analysis tasks like Crowd Counting, Crowd Density Estimation, Scene Understanding, Crowd Tracking, and Anomaly Detection. However, the challenge lies in handling the variations of scale, perspective, and occlusion in the crowded scene images.

Existing approaches for crowd counting can be broadly divided into Detection-based [1], Regression-based [2], and Density estimation based [3]. Earlier works on crowd counting were targeted to extract low-level features using hand-crafted representation and mapping these features to crowd count and crowd density with different regression techniques. Detection-based approaches usually work under the assumption that every person in the crowd can be detected using some sliding window based detection algorithms. But, these methods are computationally expensive and often fail to account for the person occlusions and background clutter in densely crowded images. To overcome these problems, researchers tried to map the features extracted from image patches into crowd counts using regression approaches. Different hand-crafted features like Histogram of Oriented Gradient (HOG), Local Binary Pattern (LBP), etc. have been exploited to represent the crowd density. However, these features fail to capture the range of variations in scale, viewpoint, and scene. Lempitsky et al. [4] proposed a linear mapping of local patch features with the corresponding density maps. By integrating over the density maps, we can estimate the crowd count. Pham et al. [5] used a non-linear mapping using random forest regression. These regression-based approaches estimated the global crowd count but failed to capture the image's spatial information.

Convolutional Neural Network (CNN) has been successful in various computer vision tasks because of its superior representation learning ability. Various CNN-based approaches have been proposed for crowd counting as well. Wang et al. [6] introduced CNN approaches to estimate crowd density through an end-to-end regression model of deep CNN. Walach et al. [7] used CNN with layered boosting structure. C. Zhang et al. [8] proposed a CNN to learn the crowd count and the crowd density alternatively. Y. Zhang et al. [9] used a multi-column CNN to extract features at multiple scales. Sam et al. [10] proposed Switch-CNN to train a classifier to choose the optimal regressor for the given image patch from multiple independent regressors. Li et al. [11] proposed CSRNet to aggregate multi-scale contextual information by using dilated convolution layers. However, most of these methods are not robust to scale, viewpoint and perspective variations.

In this paper, we propose multi-task learning [12] based deep architecture to implicitly capture the high-level scale information while generating the density map using dilated CNN. Effectively, the Crowd-Count Group Classification encodes the various scale patterns that often appear in images. This encoded contextual information in the extracted features results in generation of better density map.

---
[*], [♦] equal contribution

## 2. PROPOSED METHOD

The proposed network MTCNet (Figure 1) is deep-cascaded convolutional neural networks for two related tasks: Crowd Density Estimation (main task) and Crowd-Count Group Classification (auxiliary task). Both tasks share the same frontend, which consists of the first ten layers from the VGG-16 [13] network. The input to the MTCNet network are images of flexible size and resolution. The features extracted from frontend are fed to each task. Figure 2 shows the end-to-end architecture of the proposed network. The hyper-parameters of the main task are similar to the values used in [11]. The output features from intermediate CNN blocks (as shown below) are $X_1$ and $X_2$ respectively. Both $X_1$ (512 feature maps) and $X_2$ (128 feature maps) are concatenated, and the resulting 640 feature maps are fed to the "Crowd Density Estimation" block, which is dilated CNN network with dilation rate of two. In the auxiliary task, only $X_2$ is fed to the ten-class CNN classifier network.

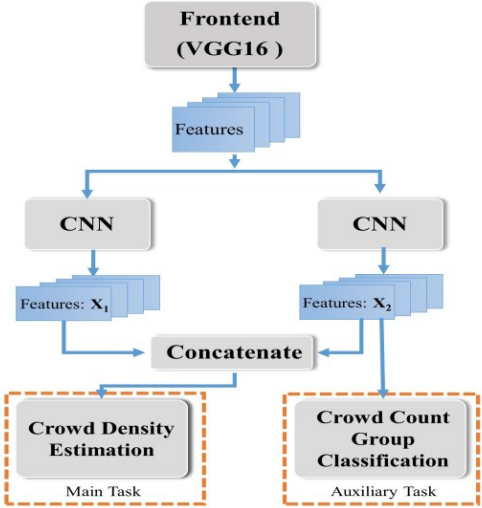

Figure 1: High-level architecture of the proposed MTCNet

### 2.1. Dilated Convolution

Pooling layer is frequently used to avoid overfitting, but at the same time, it results in loss of spatial information of the features. One way to recover back the lost information is to use deconvolutional layers. However, these further increases the computation and are hard to train. To alleviate both these shortcomings, we use Dilated Convolution [14]. Its use of sparse kernels increases the receptive field without any computation overhead. In dilated convolution with $d$ dilation stride, a standard kernel size of $k$ x $k$ filter is expanded to $k + (k-1)$ x $(d-1)$ which allows flexibility to capture contextual information on multiple scales while maintaining the same resolution. 2D-Dilated convolution can be defined as –

$$y(m,n) = \sum_{i=1}^{M}\sum_{j=1}^{N} x(m+d\times i,\ n+d\times j)w(i,j) \quad (1)$$

Li et al. [11] have shown the efficiency of using dilation rate as two in comparison with other dilation rates. Thus, we decided to use dilation rate of two in our proposed MTCNet.

### 2.2. MTL Paradigm: Role of Auxiliary Task

The MTL [12] paradigm is based on training multiple tasks simultaneously by learning the commonalities and differences across the tasks and thus leads to improved performance of task-specific models as compared to training the tasks separately. To improve the performance of the main task in MTCNet, we exploit MTL by augmenting Crowd-Count Group Estimation as the auxiliary task. Both tasks are learned jointly and use shared representation. Hence, this enhances the generalization by using the domain information from each task as an inductive bias [12]. MTL also works as a regularization technique because it makes the model perform on multiple tasks. MTL regularization is superior to regularizations that prevent overfitting as it penalizes all complexity uniformly [15]. The auxiliary task can distinguish the features based on different crowd density distributions, as it aims to classify crowd count range. These features learned from the auxiliary task ($X_2$) helps the main task to better distinguish various scales of crowd densities.

### 2.3. Ground Truth Generation

The ground truth provided with the dataset only mentions the positions of people head. These given head annotations are blurred with normalized Gaussian kernel to generate the density map. For highly congested scenes, we use the geometry-adaptive kernel, defined as:

$$F(x) = \sum_{i=1}^{N} \delta(x-x_i) \times G_{\sigma_i}(x),\ \ with\ \ \sigma_i = \beta \overline{d_i} \quad (2)$$

where $x_i$ is the position of $i^{th}$ head in ground truth $\delta$, and $\overline{d_i}$ denotes the mean distance of $k$ nearest neighbors. We convolve a Gaussian Kernel $G_{\sigma_i}(x)$ with the ground truth $\delta(x-x_i)$ to generate density map.

To be consistent with the contemporary research works, we have followed the configuration ($\beta = 0.3$ and $k=3$) for ground truth generation as mentioned in [11]. For ShanghaiTech Part_A [9] and UCF_CC_50 [16] datasets, we have used geometry-adaptive kernels whereas fixed kernel with $\sigma = 3$ is used for ShanghaiTech Part_B.

For the auxiliary task, we have generated the ground truth class label by dividing the crowd-count range into ten groups of equal size as shown in Eq. 3.

$$Class\ Label = \min\left(round\left(\frac{C_i^{GT} - C^{Min}}{C^{Max} - C^{Min}} * 10\right), 9\right) \quad (3)$$

where $C_i^{GT}$ is the ground truth crowd count of an image, $C^{max}$ and $C^{min}$ are the maximum and minimum crowd count in training dataset.

## 2.4 Training Details and Loss Function

Our proposed network is an end-to-end trainable network with combined optimization of Mean Square Error (MSE) loss for the main task and Cross-Entropy (CE) loss for the auxiliary task. The first ten layers from frontend are fine-tuned from pre-trained VGG-16. We use Gaussian distribution with 0.01 standard deviation to initialize the remaining layers. We use stochastic gradient descent with a fixed learning rate at 1e-7 during training.

| Input image (variable resolution color image) ||
|---|---|
| 64 x Conv2d (3, 1) <br> 64 x Conv2d (3, 1) ||
| MaxPool2d (kernel= (2,2)) ||
| 128 x Conv2d (3, 1) <br> 128 x Conv2d (3, 1) ||
| MaxPool2d (kernel= (2,2)) ||
| 256 x Conv2d (3, 1) <br> 256 x Conv2d (3, 1) <br> 256 x Conv2d (3, 1) ||
| MaxPool2d (kernel= (2,2)) ||
| 512 x Conv2d (3, 1) <br> 512 x Conv2d (3, 1) <br> 512 x Conv2d (3, 1) ||
| 512 x Conv2d (3, 1) <br> 512 x Conv2d (3, 1) : $X_1$ | 512 x Conv2d (3, 1) <br> 128 x Conv2d (3, 1) : $X_2$ |
| Concatenation ($X_1$,$X_2$) | AdaptiveMaxPool2d (size= (64,64)) |
| 512 x Conv2d (3, 2) <br> 512 x Conv2d (3, 2) <br> 512 x Conv2d (3, 2) <br> 256 x Conv2d (3, 2) <br> 128 x Conv2d (3, 2) <br> 64 x Conv2d (3, 2) | Dense (output=512) <br> Dense (output=256) <br> Dense (output=10) |
| 1 x Conv2d (1, 1) <br> **Density Map** | Softmax <br> **Count Group Classification** |

Figure 2: Detailed architecture of the proposed MTCNet

In the main task, we use MSE loss to measure the difference between the generated and ground-truth density maps. The MSE loss function is formulated as:

$$L_1(\theta) = \frac{1}{2N} \sum_{i=1}^{N} \|D(X_i, \theta) - D_i\|_2^2 \quad (4)$$

Here, $N$ represents the training batch size and $\theta$ is the model parameters. $D(X_i, \theta)$ is the generated density map for an image $X_i$ with $D_i$ as the ground truth density map. In the auxiliary task, we use cross-entropy loss to measure count-group classification error. The CE loss can be written as:

$$L_2(\theta) = -\sum_{i=1}^{N} \sum_{c=1}^{M} y_{i,c} \log(p_{i,c}) \quad (5)$$

where, $y_{i,c}$ is a binary indicator if the true class label of $i^{th}$ training example is class $c$ and $p_{i,c}$ denotes its predicted probability. In the multi-task learning setting, the combined loss is calculated as:

$$L_{total}(\theta) = L_1(\theta) + \lambda L_2(\theta) \quad (6)$$

We experimented with different values of $\lambda$, weight factor, to get the weighted total loss to be back-propagated through the network. We have presented a comparative study for the same in the later section.

## 3. DATASET

### 3.1. ShanghaiTech dataset

The ShanghaiTech dataset [9] comprises 1198 annotated images with total 330,165 people in them. This dataset consists of two parts- Part A and Part B. Part A contains 482 randomly crawled images from internet and have high crowd density. Part B comprises 716 images and has relatively sparse crowd density. The train-test split is provided with the dataset. Part A and Part B have 300 and 400 training images respectively. The testing set contains 182 images in Part A and 316 images in Part B. The crowd count ranges from 33 to 3139 and 9 to 578 for part A and part B respectively.

### 3.2. UCF_CC_50 dataset

The UCF_CC_50 dataset [16] contains images with dramatically varying crowd density. There are 50 highly congested images with varying spatial resolution and background. The crowd count ranges from 94 to 4543 with mean crowd count of 1280 persons per image. Limited availability of training images and high variance in terms of crowd density and image size make it a challenging dataset. We use 5-fold cross-validation as mentioned in [16].

## 4. EXPERIMENTAL RESULTS AND DISCUSSION

We demonstrate the performance of MTCNet for crowd count estimation as compared to state-of-the-art methods. MTCNet outperforms them all on two benchmark datasets in terms of Mean Absolute Error (MAE) and Mean Squared Error (MSE).

### 4.1. Evaluation metrics

The MAE and the MSE used for evaluating and comparing MTCNet and other state-of-the-art methods are as follow:

$$MAE = \frac{1}{N} \sum_{i=1}^{N} |C_i^{EST} - C_i^{GT}| \quad (7)$$

$$MSE = \sqrt{\frac{1}{N} \sum_{i=1}^{N} |C_i^{EST} - C_i^{GT}|^2} \quad (8)$$

where, $C_i^{EST}$ and $C_i^{GT}$ are the estimated and ground truth crowd count in the given testing image and N is the total number of testing images.

### 4.2. Evaluation on ShanghaiTech dataset

The comparison of our proposed architecture with other contemporary works for ShanghaiTech dataset is shown in Table 1. Our proposed method achieves the lowest MAE and MSE value. Figure 3 shows representative examples of the ground truth and generated density maps. In Part A, we achieve 5.8% and 6.2% reduction over the second best in MAE and MSE respectively. For Part B, our proposed method lowers the MAE and MSE by 14.9% and 16.9% respectively compared to state-of-the-art results.

Table 1: Comparison on the ShanghaiTech dataset

|  | Part A | | Part B | |
| --- | --- | --- | --- | --- |
| Method | MAE | MSE | MAE | MSE |
| MCNN[9] | 110.2 | 173.2 | 26.4 | 41.3 |
| MSCNN [17] | 83.8 | 127.4 | 17.7 | 30.2 |
| Switch-CNN [10] | 90.4 | 135.0 | 21.6 | 33.4 |
| L2R (keyword) [18] | 73.6 | 112.0 | 13.7 | 21.4 |
| CSRNet [11] | 68.2 | 115.0 | 10.6 | 16.0 |
| Liu et al [19] | 67.6 | 110.6 | 10.1 | 18.8 |
| MTCNet (Proposed) | **63.7** | **103.7** | **8.6** | **13.3** |

### 4.3. Evaluation on UCF_CC_50 dataset

We have compared and presented the 5-fold cross-validation results on UCF_CC_50 dataset in Table 2.

Table 2: Comparison on the UCF CC 50 dataset

| Method | MAE | MSE |
| --- | --- | --- |
| Idrees et al. [16] | 419.5 | 541.6 |
| C. Zhang et al. [8] | 467.0 | 498.5 |
| MCNN[9] | 377.6 | 509.1 |
| MSCNN [17] | 363.7 | 468.4 |
| Switch-CNN [10] | 318.1 | 439.2 |
| L2R (keyword) [18] | 279.6 | 388.9 |
| CSRNet [11] | 266.1 | 397.5 |
| Liu et al [19] | 253.1 | 356.4 |
| MTCNet (Proposed) | **226.6** | **328.2** |

Our proposed model surpasses all other methods and achieves the lowest MAE and MSE value. It obtains 10.5% and 7.9% improvement in MAE and MSE respectively as compared to the second best method.

### 4.4. Effect of the weight factor, λ on performance

We experiment with five different values of λ (weight factor) for our model on Shanghai Dataset- Part A. Given the complexity of the dataset and varying pattern of the objective loss functions (MSE & CE), we have heuristically experimented with different values to get the optimal value of λ; the results are shown in Table 3. Our proposed model achieves lowest MAE and MSE with 1e-3 as λ.

Table 3: Comparative study on the effect of weight factor

|  | λ= 1 | λ= 1e-1 | λ= 1e-2 | **λ= 1e-3** | λ= 1e-4 |
| --- | --- | --- | --- | --- | --- |
| MAE | 68.4 | 68.0 | 65.5 | **63.7** | 66.5 |
| MSE | 109.0 | 108.4 | 105.8 | **103.7** | 104.7 |

### 4.5. Ablation study: Effect of MTL setting

We compare the accuracy of auxiliary and main tasks in standalone and MTL settings (jointly training both tasks) to establish the effectiveness of jointly learning both tasks. The classification accuracy of auxiliary task improves from 56.6% in standalone setting (only training auxiliary task) to 79.7% in MTL setting. In addition, the main task achieves lower MAE from 68.2 in standalone setting (only training the main task) to 63.7 in MTL setting.

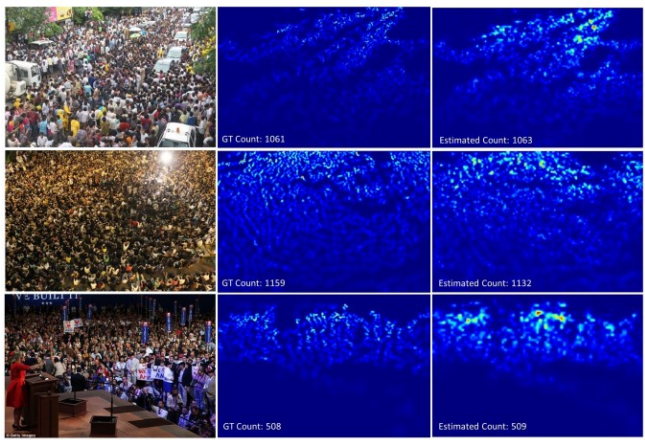

Figure 3: Representative experimental results on ShanghaiTech (Part A) dataset. Each row has the original crowd image, the ground truth and the generated density map by MTCNet.

## 5. CONCLUSION

We have proposed a novel end-to-end model, MTCNet, based on the multi-task learning paradigm to jointly train the crowd density map (main task) and the count group classification (auxiliary task). By combining two related tasks (main & auxiliary), the model is able to implicitly learn the scale factor of the given crowd scene and hence able to encode different scale features. Experimental results on benchmark datasets indicate that MTCNet attains the lowest MAE and MSE count when compared with the contemporary state-of-the-art crowd counting methods. We have also evaluated the effect of different weight factors on combined loss and presented the ablation study on the effectiveness of the auxiliary task. We will explore different network architectures for adding perspective based prior to our proposed MTCNet. We will also extend our model for vehicle counting task in the future.

## 6. REFERENCES

[1] Navneet Dalal and Bill Triggs, "Histograms of oriented gradients for human detection," in *Computer Vision and Pattern Recognition, 2005. CVPR 2005. IEEE Computer Society Conference on*. IEEE, 2005, vol. 1, pp. 886–893.

[2] Antoni B Chan and Nuno Vasconcelos, "Bayesian poisson regression for crowd counting," in *Computer Vision, 2009 IEEE 12th International Conference on*. IEEE, 2009, pp. 545–551.

[3] Bolei Xu and Guoping Qiu, "Crowd density estimation based on rich features and random projection forest," in *Applications of Computer Vision (WACV), 2016 IEEE Winter Conference on*. IEEE, 2016, pp. 1–8.

[4] Victor Lempitsky and Andrew Zisserman, "Learning to count objects in images," in *Advances in Neural Information Processing Systems*, 2010, pp. 1324–1332.

[5] V. Pham, T. Kozakaya, O. Yamaguchi and R. Okada, "COUNT Forest: CO-Voting Uncertain Number of Targets Using Random Forest for Crowd Density Estimation," *2015 IEEE International Conference on Computer Vision (ICCV)*, Santiago, 2015, pp. 3253-3261.

[6] C. Wang, H. Zhang, L. Yang, S. Liu, and X. Cao, "Deep people counting in extremely dense crowds," in *Proceedings of the 2015 ACM on Multimedia Conference*, pp. 1299–1302, 2015.

[7] E. Walach and L. Wolf, "Learning to count with cnn boosting," in *European Conference on Computer Vision*, pp. 660–676. Springer, 2016.

[8] Cong Zhang, Hongsheng Li, X. Wang and Xiaokang Yang, "Cross-scene crowd counting via deep convolutional neural networks," *2015 IEEE Conference on Computer Vision and Pattern Recognition (CVPR)*, Boston, MA, 2015, pp. 833-841.

[9] Y. Zhang, D. Zhou, S. Chen, S. Gao and Y. Ma, "Single-Image Crowd Counting via Multi-Column Convolutional Neural Network," *2016 IEEE Conference on Computer Vision and Pattern Recognition (CVPR)*, Las Vegas, NV, 2016, pp. 589-597.

[10] D. B. Sam, S. Surya and R. V. Babu, "Switching Convolutional Neural Network for Crowd Counting," *2017 IEEE Conference on Computer Vision and Pattern Recognition (CVPR)*, Honolulu, HI, 2017, pp. 4031-4039.

[11] Y. Li, X. Zhang and D. Chen, "CSRNet: Dilated Convolutional Neural Networks for Understanding the Highly Congested Scenes," *2018 IEEE/CVF Conference on Computer Vision and Pattern Recognition*, Salt Lake City, UT, 2018, pp. 1091-1100.

[12] R.Caruana, "Multitask learning," *Machine learning*, vol. *28*(1), pp.41-75, 1997

[13] K. Simonyan and A. Zisserman, "Very deep convolutional networks for large-scale image recognition," in *Proceedings of International Conference on Learning Representations*, 2015.

[14] F. Yu and V. Koltun, "Multi-scale context aggregation by dilated convolutions," in *Proceedings of International Conference on Learning Representations*, 2016.

[15] B. R. Paredes, A. Argyriou, N. Berthouze, M. Pontil, "Exploiting Unrelated Tasks in Multi-Task Learning," *Proceedings of the Fifteenth International Conference on Artificial Intelligence and Statistics*, vol. 22, pp. 951-959, 2012.

[16] H. Idrees, I. Saleemi, C. Seibert and M. Shah, "Multi-source Multi-scale Counting in Extremely Dense Crowd Images," *2013 IEEE Conference on Computer Vision and Pattern Recognition*, Portland, OR, 2013, pp. 2547-2554.

[17] L. Zeng, X. Xu, B. Cai, S. Qiu and T. Zhang, "Multi-scale convolutional neural networks for crowd counting," *2017 IEEE International Conference on Image Processing (ICIP)*, Beijing, 2017, pp. 465-469.

[18] X. Liu, J. van de Weijer and A. D. Bagdanov, "Leveraging Unlabeled Data for Crowd Counting by Learning to Rank," *2018 IEEE/CVF Conference on Computer Vision and Pattern Recognition*, Salt Lake City, UT, 2018, pp. 7661-7669.

[19] Ming Liu, Jue Jiang, Zhenqei Guo, Zenan Wang and Yang Liu, "Crowd Counting with Fully Convolutional Neural Network," *2018 25th IEEE International Conference on Image Processing (ICIP)*, Athens, 2018, pp. 953-957.